\documentclass[smallabstract,smallcaptions]{dccpaper}

\usepackage{epsfig}
\usepackage{amsmath}
\usepackage{amssymb}
\usepackage{color}
\usepackage{url}
\usepackage{booktabs}
\usepackage{marvosym}
\usepackage{ifsym}
\usepackage[nameinlink,capitalise]{cleveref}

\usepackage{booktabs}
\usepackage{multirow}

\newlength{\figurewidth}
\newlength{\smallfigurewidth}

\begin{document}

\title
{\large
\textbf{SSNVC: Single Stream Neural Video Compression with Implicit Temporal Information}
}

\author{%
Feng Wang$^{1}$, Haihang Ruan$^{2}$, Zhihuang Xie$^{3}$, Ronggang Wang$^{1, \dag}$, Xiangyu Yue$^{4}$\\[0.5em]
{\small\begin{minipage}{\linewidth}\begin{center}
\begin{tabular}{ccc}
$^{1}$Shenzhen Graduate School, Peking University, Shenzhen, China\\ 
$^{2}$Shanghai Institute of Microsystem and Information Technology, Shanghai, China\\
$^{3}$OPPO Research, Shenzhen, China \\
$^{4}$The Chinese University of Hong Kong, Hong Kong, China\\
\url{wangfeng@stu.pku.edu.cn} \  \url{&} \ \url{rgwang@pkusz.edu.cn}\\
\thanks{$^{\dag}$: Corresponding contact author email: rgwang@pkusz.edu.cn.}
\small
\end{tabular}
\end{center}\end{minipage}}
}

\maketitle
\thispagestyle{empty}

\begin{abstract}
   Recently, Neural Video Compression (NVC) techniques have achieved remarkable performance, even surpassing the best traditional lossy video codec. 
   However, most existing NVC methods heavily rely on transmitting Motion Vector (MV) to generate accurate contextual features, which has the following drawbacks. 
   (1) Compressing and transmitting MV requires specialized MV encoder and decoder, which makes modules redundant.
   (2) Due to the existence of MV Encoder-Decoder, the training strategy is complex. 
   In this paper, we present a noval Single Stream NVC framework (SSNVC), which removes complex MV Encoder-Decoder structure and uses a one-stage training strategy.
    SSNVC implicitly use temporal information 
    by adding previous entropy model feature to current entropy model and using previous two frame to generate predicted motion information at the decoder side. 
    Besides, we enhance the frame generator to generate higher quality reconstructed frame.
   Experiments demonstrate that SSNVC can achieve state-of-the-art performance on multiple benchmarks, and can greatly simplify compression process 
   as well as training process.
\end{abstract}

\Section{1 \ \ Introduction}

In recent years, deep learning-based neural video compression (NVC) schemes have achieved great success. 
The first neural video compression work DVC \cite{dvc} 
just slightly exceeded the traditional video compression standard H.264/AVC \cite{avc}. 
While the latest neural video compression work DCVC-DC \cite{dcvc-dc} has surpassed the performance of the best traditional video compression standard H.266/VVC \cite{vvc}.
This shows great potential for deep learning-based neural video compression methods.

To compress a video sequence, it is necessary to make full use of the temporal correlation information and spatial correlation information in video sequence. 
The first generation traditional video compression standard H.261 \cite{h261} suggested using the motion information to eliminate temporal redundancy information in video sequence. 
The first neural video compression work DVC \cite{dvc} also adopted the similar scheme. The specific process is as follows: the NVC method will firstly use neural image compression (NIC) method to compress the first frame of a video sequence, and the subsequent frames will use the optical flow network \cite{spynet} with the previous frame to obtain motion information, and then transmit the motion information to the decoder-side in the channel. 
After that, using the motion information to perform motion compensation on the previous decoded frame, that is, to perform an alignment operation on the previous decoded frame, which is to eliminate temporal redundant information in the video sequence. 
Then, the spatial redundant information in the video sequence will be further eliminated by residual coding \cite{dvc} or conditional coding \cite{dcvc}, and finally a frame generator is used to generate a reconstructed frame at the decoder side, which will be also used to compress the subsequent video frames.

\begin{figure}[t]
   \centering
   \includegraphics[width=0.8\linewidth]{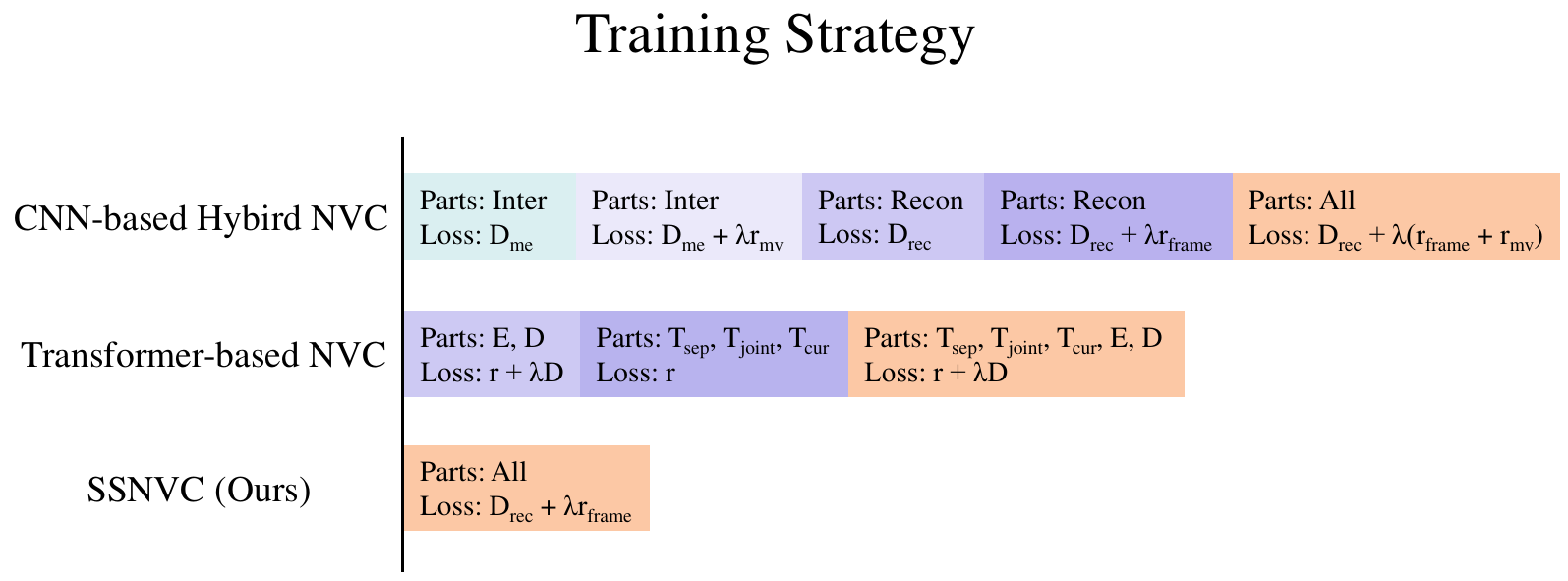}
   \vspace{-0.3cm}
   \caption{Training strategy of different neural video compression models. Each color block represents a training stage
   \cite{tcm, vct}.}
   \vspace{-0.3cm}
\label{fig:training strategy}
\end{figure}

\begin{table}[ht]
    \centering
    \vspace{-0.3cm}
    \caption{
    Using different training strategy to train DCVC-TCM \cite{tcm} model. 
    The compression ratio is measured by BD-Rate, where \textbf{positive numbers} indicate bitrate increase (\textbf{poor performance}).
    B, C, D and E represent HEVC Class B, C, D and E test datasets.}
    \vspace{0.3cm}
    \scalebox{0.8}{
    \renewcommand{\arraystretch}{1.25}
    \begin{tabular}{ccccccc}
    \toprule[1.0pt]
    Training Strategy    \ \    & B       & C       & D        & E          & Average \\ \hline
    Multi-stage          \ \    & 0.0     & 0.0     & 0.0      & 0.0        & 0.0     \\ \hline
    Single-stage         \ \    & 4.7     & 2.7     & 2.2      & 4.5        & 3.5     \\ 
    \bottomrule[1.0pt]
    \end{tabular}}
    \vspace{-0.1cm}
    \label{training strategy}
\end{table}

Most NVC methods adopt a hybrid coding framework that transmits motion information and 
spatial information (residual or conditional information)
to get better rate-distortion performance. 
However, due to the transmission of motion information, limitations are still evident in practical applications.
Firstly, it is necessary to design a dedicated motion vector (MV) encoder and decoder for the transmission of motion information, which will increase the number of modules in the entire NVC framework and lead to module redundancy. 
Secondly, due to the existence of the MV encoder and decoder, the training of the entire NVC framework is very complicated. 
When an RD (rate-distortion) loss function is directly used to train end-to-end on all modules of the entire model, the MV encoder and decoder are not well trained, and the NVC model can not reach the upper bound of its RD performance.

As shown in Table. \ref{training strategy}, using single-stage training strategy to train a NVC model (DCVC-TCM \cite{tcm} as an example) will get a poor performance than using multi-stage training strategy to train.
However, multi-stage training strategy requires that different modules need to be trained with different loss functions in different training stages, which makes the NVC model difficult to train.

Therefore, we shift our attention to eliminating the transmission process of motion information, that is removing the MV encoder and decoder on the NVC framework. 
Without transmitting motion information, there is no need to train the MV encoder-decoder specially, and multi-stage training strategy is unnecessary.
Recent work VCT \cite{vct} shows that competitive rate-distortion performance can be achieved without transmitting motion information in the channel. 
However, 
VCT still contains three different training stages.
Different modules are trained in different stages, and different loss functions need to be used, which makes the training process complicated as well.

In this paper, we propose a single stream neural video compression framework (SSNVC) based on the CNN architecture with implicit temporal information. 
The main difference between SSNVC and most existing NVC models is that SSNVC does not need to transmit the motion information in the channel for simpler compression process and training process.
As shown in Fig. \ref{fig:training strategy}, the entire training process only needs to use one loss function to perform end-to-end training on all modules in the framework.
In order to eliminate temporal redundancy, SSNVC implicitly use temporal information by adding previous entropy model feature to current entropy model and using previous two frames to generate predicted motion information at the decoder side.
Our method owns competitive rate-distortion performance compared with NVC methods with transmitting motion information.

\begin{figure*}[t]
    \centering
    \includegraphics[width=0.96\linewidth]{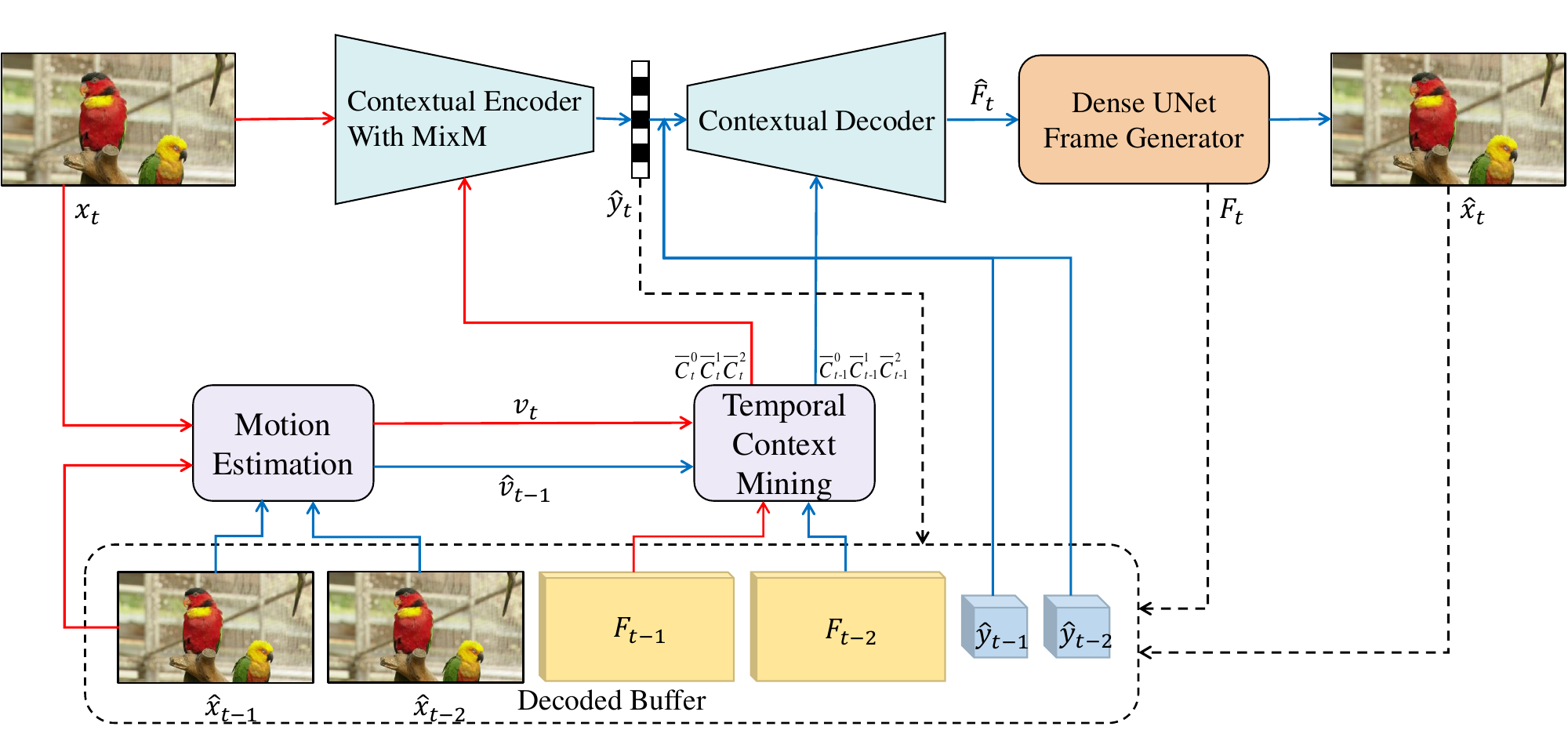}
    \caption{Overview of our proposed video compression scheme. The red solid lines are only used at the encoder side. The blue solid lines are only used at decoder side. }
    \label{fig:framework}
    \vspace{-0.2cm}
\end{figure*}

Our key contributions can be summarized as follows:
\begin{itemize}
    \item To the best of our knowledge, SSNVC is the first NVC framework based on the CNN architecture that does not transmit motion information. 
    This design can significantly simplify compression process as well as training process.
    \item SSNVC enhances frame generator and implicitly utilizes temporal information by adding previous entropy model feature to current entropy model and using previous two frames to generate predicted motion information at decoder side.
    \item SSNVC can achieve state-of-the-art performance on multiple benchmarks compared to 
    the neural video compression methods with MV encoder-decoder. 
\end{itemize}

\Section{2 \ \ Analysis of Neural Video Compression}
At the very core of contemporary NVC models lie the removal of redundancy among video sequences 
(residual coding or conditional coding).
Since two adjacent frames in a video sequence have a very high similarity, only transmitting the difference between the two frames can save a lot of bit-streams. 
This mode can be divided into two categories, residual coding and conditional coding. 
Among them, the methods of residual coding such as DVC \cite{dvc} and DVCPro \cite{dvcpro} perform residual compression on the motion compensated image directly. 
The conditional coding such as DCVC \cite{dcvc} and DCVC-TCM \cite{tcm} apply the motion information to the conditional encoder and conditional decoder. 
We use conditional coding for the spatial redundancy removal module because the entropy of residue coding is greater than or equal to that of conditional coding \cite{dcvc}.
Besides, we add mixing global and local context module (MixM) \cite{mixlic} into conditional coding for capturing both global and local dependencies to further eliminate spatial redundancy.

Although all high-performance CNN-based NVC models will transmit motion information in the channel to obtain better temporal information at the decoder side. 
Nevertheless, in most cases, it is the transmission of motion information that complicates the overall model of NVC and increases the difficulty of model training. 
Although we have removed the encoder and decoder of the motion information, the conditional encoder in the encoder side will still use the motion information between the current frame and the previous frame (but they do not need to be transmitted to the decoder side), and the conditional decoder uses the motion information between the previous decoded frame and 
the frame before previous decoded frame
at decoder side. 
At the same time, 
we add previous entropy model feature to current entropy model for accurately predicting the probability distribution of the quantized latent representation.
By utilizing the above two ways, SSNVC can implicitly use temporal information without transmitting the motion information in the channel.

\Section{3 \ \ Method}

\SubSection{3.1 \ Model Overview}
Aiming to find a succinct CNN-based neural video compression architecture, we propose SSNVC model, which eliminates the transmission process of motion information. 
This can bring two main benefits.
First, only single bit-stream needs to be transmitted in the channel, that is, the latent representation of current frame to be encoded, and there is no need to transmit motion information. 
Second, since the network does not contain MV encoder-decoder, only a single-stage training strategy is required to complete model training while achieve excellent 
rate-distortion performance. 

An overview of our scheme is depicted in \cref{fig:framework}. 
In addition to redesigning the NVC framework, we also apply a better window-based attention 
NIC method with channel-wise and checkerboard auto-regression entropy model for intra-frame compression, 
implicitly use temporal information
and enhance frame generator for stronger generation capabilities.

\textbf{Intra-frame Image Compression.} 
Better I-frame reconstruction quality is beneficial to improve the P-frame reconstruction quality of subsequent video frames. 
Inspired by \cite{stf}, we add Window-based Attention at the main encoder and main decoder of the Intra-frame image compression module. 
At the entropy model, we use a mean-scale Gaussian entropy model with channel-wise and checkerboard auto-regression \cite{elic} to build a parallelization-friendly decoder.

\textbf{Motion Estimation.} 
Our framework aims to weaken the expressiveness for motion information coding, so we use the SPyNet \cite{spynet} to estimate the MV $v_t$ between the adjacent frames. 
Besides, due to the lightweight of SpyNet, using it can reduce the complexity and improve the inference speed of SSNVC model.

\textbf{Contextual Encoder-Decoder.}
DCVC \cite{dcvc} proved that conditional coding is superior to residual coding in the field of neural video compression, because the entropy of residue coding is greater than or equal to that of conditional coding. 
Follow DCVC \cite{dcvc} and its improved work DCVC-TCM \cite{tcm}, we also adopt the conditional coding structure to design the contextual encoder-decoder. 
Unlike DCVC and DCVC-TCM, which use the current decoded MV $\hat{v}_{t}$ at both contextual encoder and contextual decoder to generate temporal conditional priors, our SSNVC uses unencoded MV $v_t$ to generate temporal conditional priors at contextual encoder and MV $\hat{v}_{t-1}$ at contextual decoder.

Besides, we add mixing global and local context module (MixM) \cite{mixlic} into contextual encoder to capture both global and local dependencies.
We use the asymmetric contextual encoder and decoder structure, which not only can obtain a more accurate latent representation, but also maintain low complexity.
We use this redesigned Contextual Encoder-Decoder to compress the current frame $x_t$ and generate reconstruction feature $\hat{F}_{t}$.
Then feeding this reconstruction feature into Dense-UNet Frame Generator to generate final reconstructed frame $\hat{x}_{t}$.

\begin{figure*}[t]
  \centering
  \begin{minipage}[c]{0.5\linewidth}
  \centering
  \includegraphics[width=0.86\linewidth]{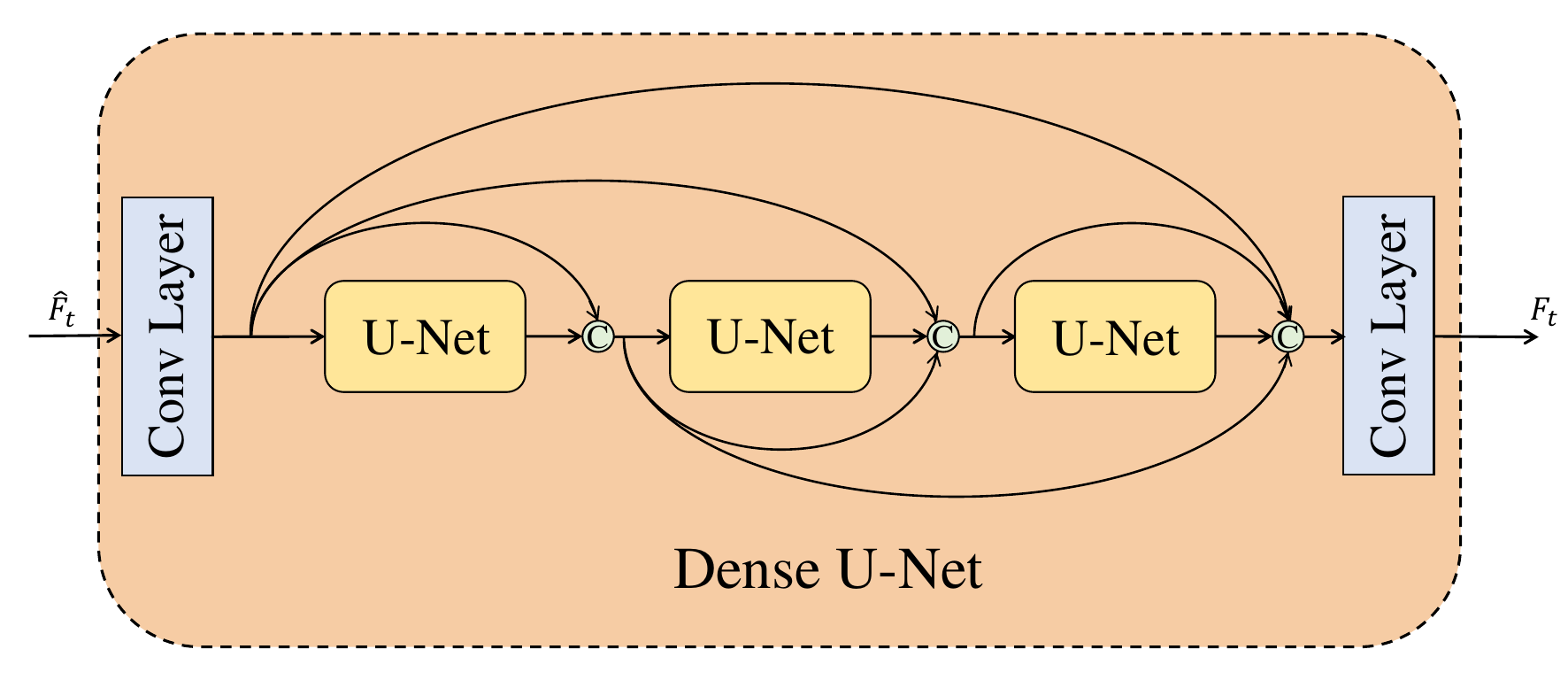}
  \end{minipage}%
  \begin{minipage}[c]{0.45\linewidth}
  \centering
  \includegraphics[width=0.86\linewidth]{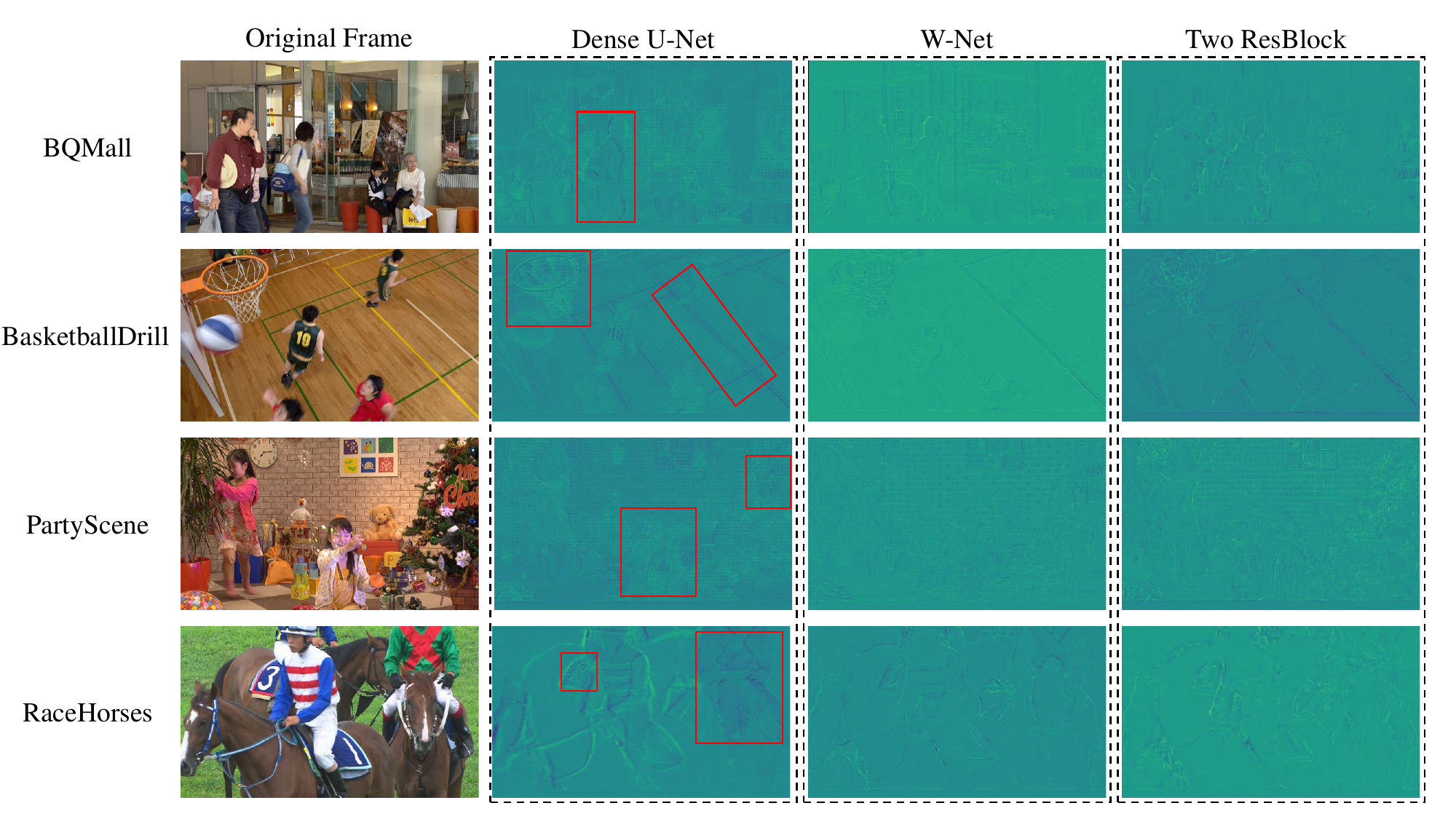}
  \end{minipage}%
  \caption{\textbf{Left:} Structure of Dense U-Net module. \textcircled{c} stands for concat. \textbf{Right:} Visualization of the features before the last convolutional layer of frame generator.
  Pictures from HEVC Class C.}
  \label{dense unet}
  \vspace{-0.45cm}
\end{figure*}

\textbf{Dense-UNet Frame Generator.}
The frame generator is introduced to generate the high-quality reconstructed frame $\hat{x}_{t}$. 
Inspired by DCVC-HEM \cite{dmc}, we use U-Nets to design our frame generator. 
Considering the complexity of the whole model, DCVC-HEM just use two cascaded U-Nets. 
This is because DCVC-HEM uses some network parameters to design MV Encoder and MV Deocder in the 
MV
coding part.
And DCVC-HEM neglects to fuse features of different dimensions in frame generator. 
Fusing features of different dimensions does not bring too much increase in model complexity, but it can significantly improve the generation ability of frame generator which will improve the compression performance of NVC model.

To improve the reconstruction ability (or generation ability) of frame generator, 
we propose a Dense U-Net frame generator to generate the high-quality reconstructed frame $\hat{x}_t$.
As shown in Fig. \ref{dense unet} (Left), 
we use three cascaded U-Nets to further enlarge the receptive field compared with DCVC-HEM which uses two U-Nets (W-Net),
and perform dense connection among the three U-Nets to fully fuse features of different dimensions.
We visualize the features before the last convolutional layer in three different frame generators, which is two ResBlock \cite{tcm}, W-Net \cite{dmc} and Dense U-Net (Ours).
From Fig. \ref{dense unet} (Right), we can find that the feature in Dense U-Net frame generator contain more information and finer textures.

\textbf{Implicit Temporal Information.} 
In order to eliminate temporal redundancy without transmitting motion information in the channel,
SSNVC implicitly use temporal information.  
Specifically, we add previous entropy model feature to current entropy model and use previous two frames to generate predicted motion information at the decoder side, as shown in Fig.  \ref{fig:framework} with the blue solid lines.

Due to the strong temporal correlation between frames in the video sequence, the motion information between two adjacent frames in the video sequence also has a certain similarity or repeatability. 
This similarity or repeatability is even higher especially in scenes with slow motion.
Therefore, the motion information between previous frames can be used to predict motion information between current frames.
At the decoder side, we use MV $\hat{v}_{t-1}$ as the predicted value of MV $\hat{v}_{t}$,
guiding Temporal Context Mining module to generate multi-scale temporal context conditional prior.

Entropy model of the contextual decoder utilizes temporal correlations between latent representations to enhance the accuracy of predicting the probability distribution of quantized latent representations
Inspired by DCVC-HEM, we use quantized latent representation of the previous frames to predict the distribution of that in the current frame. 
The latent prior explores the temporal correlation of the latent representation across frames.
Different from DCVC-HEM, SSNVC's quantized latent representation of the decoder side is completely generated at the decoder, does not rely on the encoder, and only acts on the decoder.
This ensures that temporal information does not need to be explicitly transmitted in the channel, maintaining the single stream characteristic of SSNVC. 
In addition, we also feed previous two quantized latent representations into entropy model of contextual decoder instead of previous one, in order to make full use of the temporal correlation of latent representations.

Through the above two methods, SSNVC can implicitly utilize the temporal information between different dimensions and different features in video sequences, reducing the temporal redundancy among video sequences without transmitting motion information (temporal information).

\SubSection{3.2 \ Loss Function and Training Strategy}

Our method aims to jointly train all network modules using a single loss function. We use a single-stage training strategy to optimize the 
rate-distortion (R-D) cost.
The loss function is as follow,
\begin{equation}
L=\lambda D + R= \lambda d(x_t,\hat{x}_t) +R_{\hat{f}}
\vspace{-0.1cm}
\label{simplenvc loss}
\end{equation}

Prior to this, most neural video compression methods \cite{dvc, dvcpro, dcvc, mlvc, rlvc, tcm, dmc} usually adopted a loss function or its variants as shown below,
\begin{equation}
\vspace{-0.1cm}
L=\lambda D + R= \lambda d(x_t,\hat{x}_t) + R_{\hat{v}} +R_{\hat{f}}
\label{tcm loss}
\end{equation}
where $d(x_t,\hat{x}_t)$ refers to the distortion between the input frame $x_t$ and reconstructed frame $\hat{x}_t$, where $d(\cdot)$ denotes the MSE 
or 1$-$MS-SSIM. $R_{\hat{v}}$ represents the bit rate used for encoding the quantized motion vector latent representation and the associated hyper prior. $R_{\hat{f}}$ represents the bit rate used for encoding the quantized contextual latent representation and the associated hyper prior. 
Obviously, our loss function does not include motion vector related parts.
This is the advance of our method.

Since our SSNVC model removed the MV Encoder and MV Decoder modules, it does not need to supervise the reconstruction of motion vector like previous models, and does not need to transmit motion information related bit-streams in the channel.

Benefit from this, in the training stage shown in \cref{fig:training strategy}, all training stages involving motion information can be cancelled. 
Therefore, we only need one training stage to complete the training of entire model, which greatly simplifies the training process of NVC model. 
Simplifying the training process is one of the main contributions of our work. 
However, for better enabling the model to have better inter-frame coding capabilities and dealing with the error propagation problem, we need to change the number of video frames required for training during the training process.

\Section{4 \ \ Experiments}

\SubSection{4.1 \ Experimental Setup}

\textbf{Dataset.} Vimeo-90k \cite{vimeo90k} is used for training, and HEVC Class B (1080P), C (480P), D (240P), E (720P) \cite{hevc}, UVG (1080P) \cite{uvg} and MCL-JCV (1080P) \cite{mcl-jcv} for testing.
When training, the Vimeo-90k video sequences will be randomly cropped into 256x256 patches for less memory usage and faster training speed.

\textbf{Evaluation Metrics.}  We use PSNR (Peak Signal to Noise Ratio) and MS-SSIM to measure the quality of the reconstructed frames in comparison to the original frames. And using BPP (Bit Percent Pixel) to measure the number of bits for encoding the 
contextual latent representation only.

\textbf{Implementation Detail.}
Different from some existing NVC models \cite{mlvc, dcvc, tcm, dmc} that use a multi-stage training strategy, SSNVC uses a single-stage training strategy for easier training. 
We only use one loss functions Equation $(\ref{simplenvc loss})$ to train all modules.

For experiments, we use AdamW optimizer and set batch to 4. For learning rate, it is set as 1e-4 at start and 1e-5 for fine-tuning. 
For comparing 
with other methods, we follow DCVC-TCM\cite{tcm} and train four models with different $\lambda$ values (256, 512, 1024, 2048) for multiple coding rates. 
When using MS-SSIM for performance evaluation, each $\lambda$ value is divided by 50 and using 1 - MS-SSIM as distortion loss for fine-tuning.

\SubSection{4.2 \ Experimental Results}

To demonstrate the advantage of SSNVC, we compare with the existing traditional video codec including HM \cite{HM} and VTM \cite{VTM} 
with the highest-compression-ratio settings for low-delay coding.
We also compare with previous NVC methods including DVCPro \cite{dvcpro}, MLVC \cite{mlvc}, RLVC \cite{rlvc}, DCVC \cite{dcvc} and DCVC-TCM \cite{tcm}.

\begin{figure*}[!ht]
  \centering
  \begin{minipage}[c]{0.33\linewidth}
  \centering
  \includegraphics[width=\linewidth]{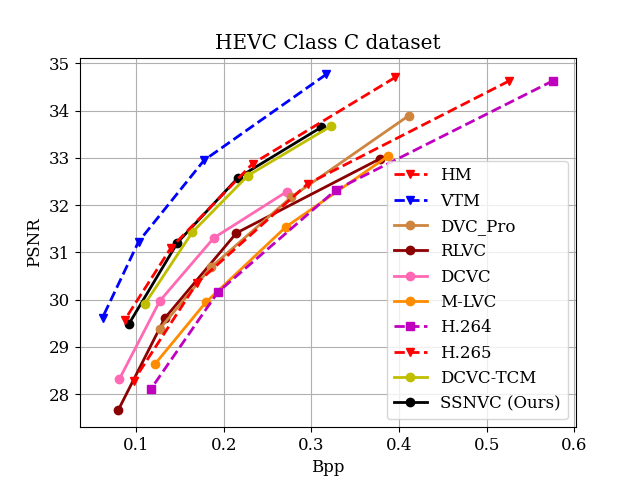}
 \end{minipage}%
  \begin{minipage}[c]{0.33\linewidth}
  \centering
    \includegraphics[width=\linewidth]{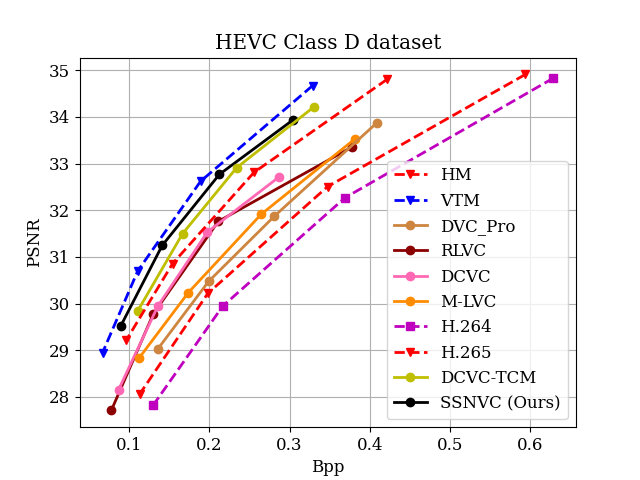}
  \end{minipage}%
  \begin{minipage}[c]{0.33\linewidth}
  \centering
    \includegraphics[width=\linewidth]{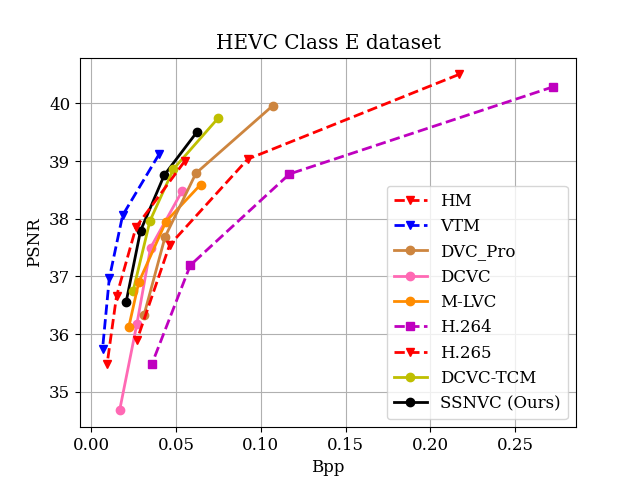}
  \end{minipage}%
    \vspace{-0.1cm}
    \caption{Rate-distortion performance of SSNVC on the HEVC Class C, D and E datasets.}
  \label{simplenvc_result}
  \vspace{-0.2cm}
\end{figure*}

\begin{table*}[ht]
    \centering
    \vspace{-0.2cm}
    \caption{BD-Rate  (\%) comparison for PSNR. The anchor is HM-16.2.}
    \vspace{-0.2cm}
    \scalebox{0.8}{
    \renewcommand{\arraystretch}{0.9}
    \small
      \begin{tabular}{ccccccccc}
      \toprule[1.0pt]
                              & B & C  & D  & E & UVG    & MCL-JCV  & HEVC Avg & All Avg  \\ \hline
      HM-16.2  \cite{HM}      & 0.0    & 0.0     & 0.0    & 0.0     & 0.0    & 0.0      & 0.0          & 0.0          \\ \hline
      VTM-13.2 \cite{VTM}     & -16.8  & -11.7   & -10.4  & -18.3   & -7.6   & -13.3    & -14.3        & -13.0   \\ \hline
      DVCPro   \cite{dvcpro}  & 36.9   & 46.3    & 30.9   & 79.2    & 43.5   & 38.7     & 48.3         & 45.9    \\ \hline
      RLVC     \cite{rlvc}    & 54.0   & 58.9    & 35.4   & 60.4    & 63.2   & 73.6     & 52.2         & 57.6    \\ \hline
      MLVC     \cite{mlvc}    & 34.6   & 78.1    & 61.1   & 54.7    & 37.8   & 46.6     & 57.1         & 52.2    \\ \hline
      DCVC     \cite{dcvc}    & 9.5    & 26.1    & 12.7   & 33.8    & 13.8   & 10.0     & 20.5         & 17.7    \\ \hline
      DCVC-TCM    \cite{tcm}     & -15.2  & 4.7     & -10.2  & -6.4    & -21.0  & -12.5    & -6.7         & -10.1   \\ \hline
      SSNVC (Ours)        & -9.8   & -0.1    & -17.2  & -15.3   & -13.6  & 1.3      & -10.6        & -9.1     \\
      \bottomrule[1.0pt]
      \end{tabular}}
      \vspace{-0.1cm}
    \label{tab_simplenvc_psnr}
\end{table*}

\begin{table*}[!ht]
    \centering
    \vspace{-0.2cm}
    \caption{BD-Rate  (\%) comparison for MS-SSIM. The anchor is HM-16.2.}
    \vspace{-0.2cm}
    \scalebox{0.8}{
    \renewcommand{\arraystretch}{0.9}
    \small
      \begin{tabular}{ccccccccc}
      \toprule[1.0pt]
                              & B & C  & D  & E & UVG    & MCL-JCV  & HEVC Avg & All Avg  \\ \hline
      HM-16.2  \cite{HM}      & 0.0    & 0.0     & 0.0    & 0.0     & 0.0    & 0.0      & 0.0          & 0.0     \\ \hline
      VTM-13.2 \cite{VTM}     & -13.0  & -11.9   & -9.6   & -14.5   & -9.3   & -12.3    & -12.3        & -11.7   \\ \hline
      DVCPro   \cite{dvcpro}  & -14.7  & -17.5   & -31.4  & 5.0     & -2.9   & -17.0    & -14.7        & -13.1   \\ \hline
      RLVC     \cite{rlvc}    & -3.7   & -4.4    & -24.7  & -5.3    & 16.0   & 10.5     & -9.5         & 1.9     \\ \hline
      MLVC     \cite{mlvc}    & 27.0   & 27.3    & 21.2   & 38.8    & 39.4   & 33.6     & 28.6         & 31.2    \\ \hline
      DCVC     \cite{dcvc}    & -24.2  & -29.8   & -39.7  & -19.5   & -16.1  & -30.7    & -28.3        & -26.7   \\ \hline
      DCVC-TCM    \cite{tcm}     & -48.5  & -47.2   & -55.1  & -53.9   & -35.0  & -44.4    & -51.2        & -47.3       \\ \hline
      SSNVC (Ours)        & -44.3      & -48.7       & -60.0      & -63.2       & -25.0      & -39.4        & -54.0            & -46.7       \\
      \bottomrule[1.0pt]
      \end{tabular}}
      \vspace{-0.4cm}
    \label{tab_simplenvc_msssim}
\end{table*}

Table. \ref{tab_simplenvc_psnr} and Table. \ref{tab_simplenvc_msssim} show the BD-Rate (\%) comparisons in terms of PSNR and MS-SSIM, respectively.
Following most existing NVC models, the intra period is set to 12.
We choose HM \cite{HM} as our anchor. Negative values indicate bit rate saving compared with HM while positive values indicate bit rate increasing.
From Table. \ref{tab_simplenvc_psnr}, we can find that SSNVC achieves an HEVC average of 10.6\% bitrate saving over HM on HEVC datasets (Class B, C, D, E) and an all average of  9.1\% bitrate saving over on all datasets. 
Compared with DCVC-TCM \cite{tcm} which is the most similar 
to SSNVC, our scheme achieves an HEVC average of 3.85\% bitrate saving in terms of PSNR on HEVC datasets.
When oriented to MS-SSIM, SSNVC can bring larger improvement. Table. \ref{tab_simplenvc_msssim} shows that it achieves an HEVC average of 54\% bitrate saving over HM on HEVC datasets (Class B, C, D, E) and an all average of  46.7\% bitrate saving over on all datasets.
We draw the RD-curves on HEVC Class C, D and E for intuitively presenting experimental results, as shown in Fig. \ref{simplenvc_result}. 
It shows that SSNVC can reconstruct higher quality decoded video frames when consumes the same bits.

\SubSection{4.3 \ Ablation Study}

\begin{table}[t]
    \centering
    \caption{Effectiveness of different components of our scheme. The compression ratio is measured by BD-Rate, where \textbf{negative numbers} indicate bitrate save (\textbf{good performance}).}
    \vspace{0.3cm}
    \scalebox{0.7}{
    \renewcommand{\arraystretch}{1.25}
    \begin{tabular}{cccccccc}
    \toprule[1.5pt]
    Win-Attn Intra & Implicit Temporal Information & Dense UNet & B & C & D & E & Average \\
    \hline
    $\times$ &$\times$ &$\times$ &  0.0  & 0.0   & 0.0   & 0.0   & 0.0   \\ \hline
    $\checkmark$ &$\times$ &$\times$   & -6.1  & -11.3 & -10.7 & -5.1  & -8.3  \\ \hline
    $\times$ &$\checkmark$ &$\times$   & -5.4  & -1.8  & -2.4  & -3.6  & -3.3  \\ \hline
    $\times$ &$\times$ &$\checkmark$   & -14.5 & -15.8 & -14.5 & -8.4  & -13.3 \\ \hline
    $\checkmark$ &$\checkmark$ &$\times$     & -13.3 & -15.2 & -14.6 & -9.2  & -13.1 \\ \hline
    $\checkmark$ &$\times$ &$\checkmark$     & -20.6 & -27.1 & -25.2 & -13.5 & -21.6 \\ \hline
    $\times$ &$\checkmark$ &$\checkmark$     & -17.9 & -15.1 & -14.6 & -10.2 & -14.5 \\ \hline
    $\checkmark$ &$\checkmark$ &$\checkmark$       & -26.9 & -30.1 & -29.7 & -18.9 & -26.4 \\ 
    \bottomrule[1.0pt]
    \end{tabular}}
    \vspace{-0.4cm}
    \label{simplenvc main ablation}
\end{table}

To verify the effectiveness of each component, we conduct comprehensive ablation studies. 
We analyze the effect of Win-Attn Intra, Implicit Temporal Information and Dense UNet.
Besides, we conduct additional experiment to prove 
the performance improvement brought by 
dense connection in frame generator.
For simplification, all experiments use HEVC testsets and the comparisons are measured by BD-Rate (\%).

\textbf{Effectiveness of Different Components.}
To demonstrate the effectiveness of different components in our scheme, we build an \textbf{anchor} scheme. 
It is based on DCVC-TCM \cite{tcm} and removes MV Encoder and MV Decoder. 
In order to use the re-filled multi-scale temporal contexts $\bar{C_t}^{l}$, the motion information $v_t$ between $x_t$ and $\hat{x}_{t-1}$ is directly used to generate $\bar{C_{t}^{l}}$ through the TCM module at the encoder side, and the motion information $\hat{v}_{t-1}$ between $\hat{x}_{t-1}$ and $\hat{x}_{t-2}$ is used to generate $\bar{C_{t-1}}^{l}$ through the same TCM module aforementioned at the decoder side.
The reason for using $\hat{v}_{t-1}$ instead of $v_t$ is that the $v_t$ can not be obtained at the decoder side (because MV is not transmitted in the channel).
Besides, the \textbf{anchor} scheme also refers to the contextual latent representation and reconstructed features of two adjacent previous frames.

To understand the contributions of our proposed 
components, 
we start with the \textbf{anchor} and gradually insert these components. 
We redesign Win-Attn Intra with channel-wise and checkerboard entropy model
to get better I-frame  reconstruction quality, which is beneficial to 
subsequent frame's reference.
We implicitly utilize temporal information between different dimensions and
different 
features in video sequences, reducing temporal redundancy among video sequences without transmitting motion information. 
Instead of residual block based frame generator which used in DCVC-TCM, We utilize
Dense UNet frame generator to generate more high-quality reconstructed frame $\hat{x}_{t}$.
From Table. \ref{simplenvc main ablation}, we can find that each component and their different combinations can improve the compression efficiency in different degrees.

\Section{5 \ \ Conclusion}
In this paper, we propose Single Stream Neural Video Compression, SSNVC. 
It implicitly utilizes temporal information to eliminate temporal redundancy in video sequence.
Without MV encoder-decoder, it only needs to transmit single bitstream in channel and use single-stage training strategy, which can greatly simplify training process and compression process of neural video compression. 
Besides, we reimplement window-based attention intra-frame image compression with channel-wise and checkerboard auto-regression entropy model, 
enhance contextual encoder with mixing global and local context module,
and redesign Dense-UNet frame generator with stronger generation capability 
to improve SSNVC's compression performance. 
Experiment results show that SSNVC can achieve 
competitive performance on multiple benchmarks.

\Section{References}
\bibliographystyle{IEEEbib}
\bibliography{refs}

\end{document}